\PassOptionsToPackage{sort}{natbib}

\documentclass[11pt]{article}

\usepackage[final]{acl}

\usepackage{times}
\usepackage{latexsym}
\usepackage[T1]{fontenc}
\usepackage[utf8]{inputenc}

\usepackage{microtype}
\usepackage{inconsolata}

\usepackage{amsmath}
\usepackage{amsfonts}
\usepackage{amssymb}

\usepackage{graphicx}
\usepackage{booktabs}
\usepackage{adjustbox}
\usepackage{geometry}
\usepackage{multirow}
\usepackage{makecell}
\usepackage{float}
\usepackage{subcaption}
\usepackage{longtable}
\usepackage{siunitx} 

\usepackage[table,dvipsnames]{xcolor} 
\usepackage{colortbl}
\definecolor{ourblue}{RGB}{219, 234, 254}  
\definecolor{loragray}{RGB}{243, 244, 246} 

\usepackage{tikz}
\usepackage{pgfplots}
\pgfplotsset{compat=1.18}

\usepackage[ruled,vlined]{algorithm2e}
\usepackage{enumitem}
\usepackage{pifont} 

\usepackage[nameinlink,capitalise]{cleveref}
\usepackage{hyperref}

\newcommand{\model}{PromptEmbedder}
\newcommand{\Memb}{\ensuremath{\mathcal{M}_{\text{pretrained}}}}
\newcommand{\Mprompt}{\ensuremath{\mathcal{M}_{\text{prompt}}}}

\renewcommand{\cite}{\citep}

\title{\model{}: Efficient and Transferable Text Embedding  via Dual-LLM Soft Prompting}

\author{
  Yu-Che Tsai\textsuperscript{1} \quad
  Kuan-Yu Chen\textsuperscript{1} \quad
  Yuan-Hao Chen\textsuperscript{1} \\[3pt]
  \textbf{Yu-Han Chang\textsuperscript{1}} \quad
  \textbf{Ching-Yu Tsai\textsuperscript{1}} \quad
  \textbf{ Yu-Hsiang Chuang\textsuperscript{1}} \quad
  \textbf{Shou-De Lin\textsuperscript{1,2}} \\
  \textsuperscript{1}Department of Computer Science and Information Engineering, National Taiwan University \\
  \textsuperscript{2}National Taiwan University AI Center of Research Excellence \\
  Taipei, Taiwan \\
  \small\texttt{\{f09922081,d13922034,b11902172,d14922008,r14922006,b12902129,sdlin\}@csie.ntu.edu.tw}
}

\begin{document}
\maketitle

\begin{abstract}
Large Language Models (LLMs) have demonstrated remarkable efficacy in text embedding, yet current adaptation methods like LoRA face significant bottlenecks in computational efficiency and cross-architecture transferability. Whenever a new backbone emerges, existing approaches require costly retraining from scratch. To address this, we propose PromptEmbedder, a novel dual-LLM framework that decouples embedding knowledge from specific backbone weights. PromptEmbedder utilizes a Prompting LLM to generate instruction-aware soft prompts for a frozen Embedding LLM via a differentiable generation process with continuous relaxation, ensuring full gradient flow during contrastive training. By localizing task-specific knowledge within the Prompting LLM, adapting to new architectures requires only retraining a lightweight linear alignment matrix. Evaluations on the MTEB benchmark show that PromptEmbedder achieves comparable performance with LoRA finetuning while reducing GPU memory by 40\% and accelerating training by 3.7$\times$. Our approach establishes a scalable, architecture-agnostic paradigm for efficient LLM-based representation learning.
\end{abstract}

\begin{figure}[t]
    \centering
    \includegraphics[width=\linewidth]{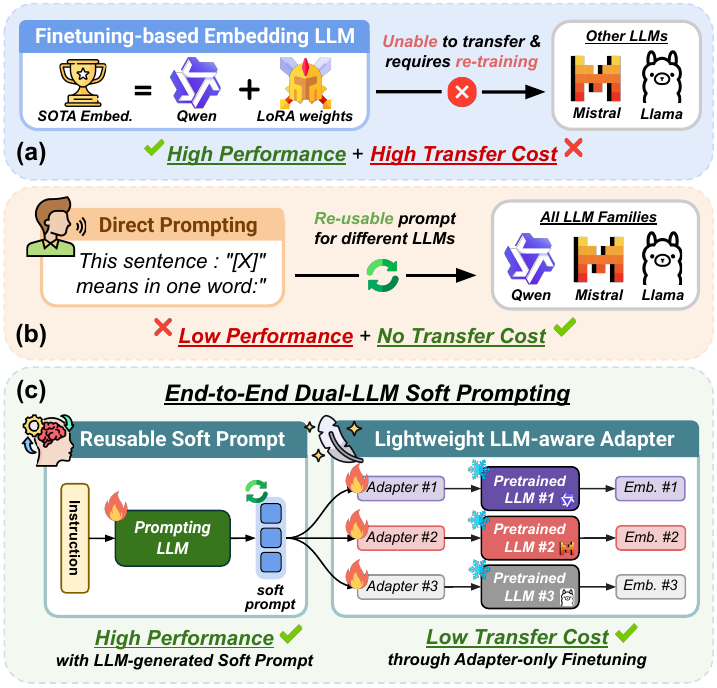}
    \caption{
Comparison of existing embedding approaches and \model{}.
(a) Fine-tuning-based methods achieve high performance but require costly retraining for each new backbone.
(b) Direct prompting methods are reusable but suffer from substantially lower embedding quality.
(c) \model{} achieves both high performance and transferability via a dual-LLM architecture.
}
    \label{fig:trailer}
\end{figure}

\section{Introduction}
The advent of Large Language Models (LLMs) has fundamentally transformed natural language processing, extending their utility beyond traditional text generation to sophisticated representation learning tasks. Recent investigations have revealed that LLMs, when appropriately adapted, can produce high-quality text embeddings that rival or surpass dedicated embedding models across diverse downstream applications, including text classification, information retrieval, and semantic similarity~\cite{karpukhin2020dense,xiong2021approximate,liu2024once}. 

\noindent The dominant paradigm for adapting LLMs to embedding tasks relies on parameter-efficient fine-tuning (PEFT) techniques. Widely used embedding models such as E5-Mistral~\cite{E5-mistral} and GRITLM~\cite{GRIT} apply Low-Rank Adaptation (LoRA)~\cite{lora} with contrastive objectives to generate high-quality text embeddings. While effective, this approach fundamentally couples all learned embedding knowledge to the weights of the target backbone. As a consequence, whenever a new LLM architecture emerges, the entire fine-tuning pipeline must be restarted from scratch. 
According to recent release histories, major LLM backbones (e.g., Qwen series~\cite{qwen2, qwen2.5, yang2025qwen3}) have been updated every six to nine months. More importantly, training a state-of-the-art embedding model such as Qwen3-Embedding requires upwards of 150 million training samples~\cite{qwen3embedding}, which could cost millions of dollars in compute~\cite{cottier2024rising}. As LLM backbones continue to evolve rapidly, it is essential to develop a more generalized paradigm that allows for the efficient transfer of embedding knowledge across model generations.

To alleviate the intensive training process, another line of research focuses on developing training-free methods via specialized prompting techniques. For instance, PromptEOL~\cite{prompteol}, MetaEOL~\cite{metaeol}, and Echo~\cite{echo} use hand-crafted instruction templates to extract semantic representations from the hidden states of pretrained LLMs. Since these prompting methods do not alter any parameters of the pretrained LLM, they are fully transferable across architectures by design. However, without appropriate adaptation, they significantly underperform fine-tuned models on standard benchmarks, making them impractical for real-world applications.

 As illustrated in Figure~\ref{fig:trailer}, existing approaches face a fundamental trade-off between performance and transfer cost. To address these limitations, we propose \model{}, a framework that achieves strong performance while remaining efficiently transferable across LLM architectures. \model{} is built on a dual-LLM architecture consisting of a \textit{Prompting LLM} and a \textit{Pretrained LLM}. The Prompting LLM autoregressively generates $k$ soft instruction tokens, which are inserted into the input text to guide the pretrained LLM in producing text representations from the last-token hidden state.
Built on this architecture, \model{} introduces two core technical strengths: (1) \textbf{Differentiable Soft Prompt Generation.} Soft prompts are generated through a continuous relaxation over the vocabulary. This allows the Prompting LLM to be jointly optimized with the frozen pretrained LLM in an end-to-end fashion without discrete sampling. (2) \textbf{Efficient Cross-Model Transfer.} All embedding-specific knowledge is localized entirely within the Prompting LLM without modifying the pretrained LLM. As a result, transferring \model{} to an unseen pretrained LLM requires retraining only a lightweight linear alignment matrix rather than repeating the full fine-tuning pipeline.

We conduct extensive evaluations on the MTEB benchmark and demonstrate that \model{} achieves three critical advantages. In terms of effectiveness, \model{} consistently outperforms existing prompting-based embedding methods such as PromptEOL and Prompt-Tuning. 
Notably, our framework achieves 96.0\%--99.8\% of the performance of LoRA finetuning by inserting only five soft tokens into the input text.
Furthermore, \model{} exhibits superior transferability by adapting to unseen LLM backbones with 3.8$\times$ faster convergence. Specifically, \model{} achieves the same performance level as LoRA training for 66 hours in only 18 hours. Finally, in terms of efficiency, maintaining a completely frozen pretrained LLM alleviates gradient computation overhead. This design reduces GPU memory usage by 36\% to 40\% and delivers a 3.7$\times$ training speedup compared to LoRA. Consequently, \model{} establishes a new paradigm for scalable and efficient LLM-based text embeddings.

\section{Related Work}

\noindent\textbf{Fine-tuning LLMs as Embedding Models.}
Fine-tuning decoder-only LLMs has become the dominant paradigm for generating high-quality text representations. E5-Mistral~\cite{E5-mistral} was among the first to demonstrate that PEFT-based fine-tuning of LLMs with contrastive objectives substantially outperforms encoder-only models. GRITLM~\cite{GRIT} further unifies generative and embedding objectives within a single model. To address the inherent limitation of unidirectional attention in decoder-only LLMs, LLM2Vec~\cite{LLM2vec} introduces bidirectional attention combined with masked next-token prediction. NV-Embed~\cite{NV-Embed} proposes a latent attention pooling layer and a two-stage training strategy to handle false negatives in non-retrieval tasks. BGE-en-ICL~\cite{bge-icl} instead leverages in-context learning within the original LLM framework to guide embedding generation. Most recently, Qwen3-Embedding~\cite{qwen3embedding} pushes performance to new heights by training on over 150 million samples.  

\noindent \textbf{Prompting LLMs as Embedding Models.} To avoid the cost of fine-tuning, a parallel line of research explores training-free methods that elicit embeddings through prompting. PromptEOL~\cite{prompteol} and MetaEOL~\cite{metaeol} design hand-crafted instruction templates that guide frozen LLMs to summarize input text, extracting semantic representations directly from hidden states without any parameter updates. While these methods eliminate training overhead, they result in substantially lower performance compared to fine-tuned methods.
Soft prompt tuning offers a middle ground by prepending learnable continuous vectors to the input and optimizing only these parameters while keeping the backbone frozen. Prefix Tuning~\cite{li2021prefix} and Prompt Tuning~\cite{prompt_tuning} demonstrate the viability of this approach on general NLP tasks, and P-Tuning v2~\cite{ptuningv2} extends it with deep, layer-wise prefixes for greater expressiveness. In the context of embedding models, soft prompts can similarly steer a frozen backbone toward the embedding objective, avoiding model modification while still benefiting from task-specific adaptation. However, the fixed set of prompt vectors limits their ability to perform diverse tasks, hindering the performance of soft prompt tuning behind conventional finetuning approaches.

\begin{figure*}[t]
    \centering
    \includegraphics[width=0.9\linewidth]{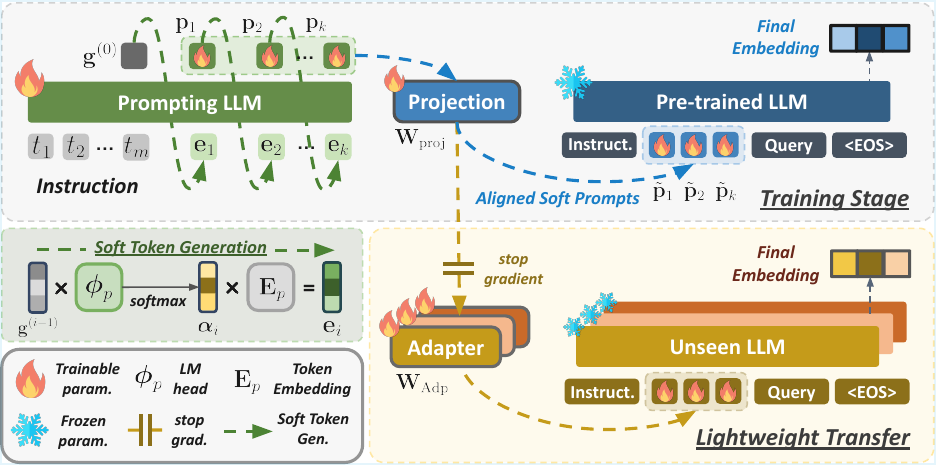}
    \caption{
Overview of the \model{} framework. \textbf{Top:} During training, \Mprompt{} generates $k$ soft prompt vectors from a task instruction, which are projected and prepended to the input of the frozen pretrained LLM \Memb{}. \textbf{Bottom-left:} Each soft token $\mathbf{e}_t$ is generated differentiably as a weighted sum over the token embedding matrix $\mathbf{E}_p$. \textbf{Bottom-right:} For lightweight transfer to an unseen backbone, only a small adapter is trained while \Mprompt{} and \Memb{} are completely frozen.
}
    \label{fig:overview}
\vspace{-15pt}
\end{figure*}

\section{Preliminary}
\paragraph{Notations.}
We consider a LLM–based embedding system that maps a text document to a dense vector representation. 
Specifically, we adopt a pretrained decoder-only language model with parameters 
$\psi = \{\mathbf{E}, \theta, \phi\}$, where 
$\mathbf{E} \in \mathbb{R}^{|\mathcal{V}| \times d}$ denotes the token embedding matrix, 
$\theta$ represents the transformer decoder parameters, and 
$\phi$ corresponds to the language modeling head used for next-token prediction. 
Here, $d$ denotes the hidden dimension and $|\mathcal{V}|$ is the vocabulary size.

\paragraph{Instruction-based Text Embedding.}
Following recent state-of-the-art embedding models~\citep{E5-mistral, GRIT, qwen3embedding}, we adopt an instruction-based embedding paradigm. 
Given a task instruction $t = [t_1, \dots, t_m]$ and an input text $X = [x_1, \dots, x_n]$, we first concatenate them into a single sequence:
\begin{equation*}
\texttt{Instruction: } t \ \texttt{Query: } X \ \texttt{<EOS>}.
\end{equation*}
Here, $t$ is a short natural-language description of the task (e.g., 
``\textit{Given a question, retrieve passages that answer the question}'' in retrieval settings).
The concatenated sequence is then mapped to token embeddings through $\mathbf{E}$ and then processed by the transformer decoder $\mathrm{Dec}_\theta$ to produce contextual hidden states. Finally, the resulting text representation is obtained through last-token pooling, which refers to the hidden state corresponding to the \texttt{<EOS>} token:
\begin{equation}
\label{eq:eos-embedding}
\mathbf{v} =
\mathrm{Dec}_\theta\!\left(
\mathbf{E}([t; X; \texttt{<EOS>}])
\right)_{\texttt{<EOS>}}
\in \mathbb{R}^{d}.
\end{equation}

Finally, the obtained embeddings are optimized by minimizing a contrastive loss via finetuning the pretrained LLM.

\section{Method}

\paragraph{Overview.}
As illustrated in ~\cref{fig:overview}, \model{} consists of a \textit{Prompting LLM} \Mprompt{} with parameters $\psi_p = \{\mathbf{E}_p, \theta_p, \phi_p\}$ and a frozen \textit{Pre-trained LLM} \Memb{} with parameters $\psi_e = \{\mathbf{E}_e, \theta_e, \phi_e\}$. Given a task description $t$, \Mprompt{} generates $k$ soft prompt vectors $\mathbf{P} = [\mathbf{p}_1, \dots, \mathbf{p}_k]$ in a fully differentiable manner. These vectors are projected into the embedding space of \Memb{} and inserted as a prefix to the input text. The final embedding is then extracted from the augmented sequence via last-token pooling. During training, only the LoRA parameters of \Mprompt{} and the projection matrix are optimized using a contrastive objective while \Memb{} remains completely frozen.

\subsection{Soft Prompt Generation through LLM}
\label{subsec:prompting_llm}
A key innovation of \model{} is to dynamically construct soft prompts from task instructions using an auxiliary LLM \Mprompt{}. Since \Mprompt{} must be trained through the frozen \Memb{}, the prompt generation process should remain fully differentiable to allow gradients from the contrastive loss to propagate back to \Mprompt{}. However, standard autoregressive generation involves discrete token sampling, which breaks the gradient flow. Inspired by prior work on soft token generation~\cite{gircse,deng2025following}, we instead generate soft prompts through an iterative process over a continuous relaxation of the vocabulary, which preserves differentiability and allows \Mprompt{} to be trained end-to-end via the embedding objective. 

\paragraph{Instruction-conditioned Soft Prompt Generation.}

Given a task instruction $t = [t_1, \dots, t_m]$, we first feed it into the prompting model \Mprompt{}, consisting of the embedding matrix $\mathbf{E}_p$ and transformer decoder $\theta_p$.
Let $\mathbf{H}^{(0)} = \mathrm{Dec}_{\theta_p}(\mathbf{E}_p(t))$ denote the final-layer hidden states of the instruction, and let
$\mathbf{g}^{(0)} = \mathbf{H}^{(0)}_{m} \in \mathbb{R}^{d_p}$
be the hidden state of the last instruction token.

Soft prompt generation then starts autoregressively from $\mathbf{g}^{(0)}$.
At generation step $i$, we first construct a soft token $\mathbf{e}_i$ from the current last hidden state $\mathbf{g}^{(i-1)}$.
Specifically, we compute a vocabulary distribution using the LM head $\phi_p$:
\begin{equation}
\boldsymbol{\alpha}_i =
\mathrm{softmax}\!\left(
\phi_p(\mathbf{g}^{(i-1)})
\right)
\in \mathbb{R}^{|\mathcal{V}|},
\end{equation}
and obtain the corresponding soft token by computing a convex combination over the vocabulary:

\begin{equation}
\label{eq:soft-token}
\mathbf{e}_i =
\boldsymbol{\alpha}_i^\top \mathbf{E}_p
\in \mathbb{R}^{d_p}.
\end{equation}

We then append $\mathbf{e}_i$ to the current sequence and pass the extended sequence through \Mprompt{} again.
Formally, let $\mathbf{Z}^{(i)} = [t; \mathbf{e}_1; \dots; \mathbf{e}_i]$
denote the instruction augmented with the first $i$ generated soft tokens.
We compute
$\mathbf{H}^{(i)} =
\mathrm{Dec}_{\theta_p}\!\left(
\mathbf{E}_p(\mathbf{Z}^{(i)})
\right)$ 
and take the hidden state of the last position as
$\mathbf{p}_i = \mathbf{H}^{(i)}_{m+i} \in \mathbb{R}^{d_p}$.
This hidden state $\mathbf{p}_i$ serves as the prompt representation retained at step $i$, while the next generation step uses $\mathbf{g}^{(i)} = \mathbf{p}_i$ as its current state.
Repeating this process for $K$ steps yields the final soft prompt sequence:
$\mathbf{P} = [\mathbf{p}_1, \dots, \mathbf{p}_K] \in \mathbb{R}^{K \times d_p}$.

\paragraph{Cross-LLM Prompt Alignment.}

Since the two pretrained models, \Mprompt{} and \Memb{}, may differ not only in hidden dimension (with hidden dimensions $d_p$ and $d_e$) but also in the structure of their embedding spaces, the generated soft prompts cannot be directly used as inputs to \Memb{}. We therefore introduce a learnable linear projection
$\mathbf{W}_{\mathrm{proj}} \in \mathbb{R}^{d_e \times d_p}$
to map each prompt vector into the embedding space of \Memb{}:

\begin{equation}
\tilde{\mathbf{p}}_i =
\mathbf{W}_{\mathrm{proj}}
\mathbf{p}_i
\in \mathbb{R}^{d_e}.
\end{equation}

This produces the aligned prompt sequence
$\tilde{\mathbf{P}} =
[\tilde{\mathbf{p}}_1, \dots, \tilde{\mathbf{p}}_k]
\in \mathbb{R}^{k \times d_e}$,
which is inserted as a prefix to the input sequence processed by \Memb{}.

\subsection{Embedding Extraction}
\label{subsec:target_llm}

Given the aligned soft prompt sequence $\tilde{\mathbf{P}}$, we insert it between the instruction $t$ and the input text $X$ to construct the input sequence for the frozen embedding model \Memb{}:
\begin{equation}
    \mathbf{X}_{\text{combined}} =
    \mathbf{E}_e(t) \Vert
    \tilde{\mathbf{P}} \Vert
    \mathbf{E}_e(X) \Vert
    \mathbf{E}_e(\texttt{<EOS>}),
\end{equation}
where $\mathbf{E}_e$ denotes the token embedding matrix of \Memb{} and $\Vert$ denotes concatenation.
The resulting sequence is processed by the frozen transformer decoder $\theta_e$ of \Memb{}, and the final text embedding is obtained by last-token pooling as described in \cref{eq:eos-embedding}:
\begin{equation}
\mathbf{v} =
\mathrm{Dec}_{\theta_e}\!\left(
\mathbf{X}_{\text{combined}}
\right)_{\texttt{<EOS>}}
\in \mathbb{R}^{d_e}.
\end{equation}

\subsection{Training Objective}
\label{subsec:training}

Following prior embedding models~\cite{E5-mistral,gircse,GRIT}, we train \model{} with a contrastive InfoNCE objective~\cite{infonce}. The training data consists of triplets $(q_i, d_i^+, d_i^-)$, where $q_i$ is a query, $d_i^+$ is a relevant document, and $d_i^-$ is a hard negative. 
For each query $q_i$, the negative set is $\mathcal{N}_i = \{d_i^-\} \cup \{d_j^+\}_{j \ne i}$, which combines the paired hard negative with in-batch positives from other examples. The loss is defined as:

\begin{equation}
\label{eq:infonce}
\mathcal{L}
=
-\frac{1}{N_b}
\sum_{i=1}^{N_b}
\log
\frac{e^{\mathrm{sim}(\mathbf{v}_{q_i},\mathbf{v}_{d_i^+})/\tau}}
{\sum_{d \in \{d_i^+\} \cup \mathcal{N}_i}
e^{\mathrm{sim}(\mathbf{v}_{q_i},\mathbf{v}_{d})/\tau}}
,
\end{equation}

where $\mathrm{sim}(\cdot,\cdot)$ is cosine similarity and $\tau=0.2$ is the temperature hyperparameter. Gradients propagate through the frozen \Memb{} to update $\mathbf{W}_{\text{proj}}$ and the LoRA parameters of \Mprompt{}, while \Memb{} remains fixed throughout training.

\subsection{Transferability to Unseen LLM Backbones}
\label{sec:transfer}

Another important advantage of \model{} is its ability to generalize across different LLM backbones without retraining the prompting model \Mprompt{} and the projection matrix $\mathbf{W}_{\text{proj}}$.
Since all task knowledge is already learned and encoded in these two modules, adapting to a new embedding model $\Memb{}'$ only requires a lightweight adapter.
Specifically, we introduce an additional trainable projection
$\mathbf{W}_{\text{adp}} \in \mathbb{R}^{d_{e'} \times d_e}$,
which is appended after the original projection layer:
\begin{equation}
\tilde{\mathbf{p}}_i' =
\mathbf{W}_{\text{adp}}
\mathbf{W}_{\text{proj}}
\mathbf{p}_i
\in \mathbb{R}^{d_{e'}}.
\end{equation}
The adapted prompt sequence
$\tilde{\mathbf{P}}' =
[\tilde{\mathbf{p}}_1', \dots, \tilde{\mathbf{p}}_k']$
is then inserted into the new embedding model $\Memb{}'$ following the same procedure as in~\cref{subsec:target_llm}. 
During the transfer learning stage, only $\mathbf{W}_{\text{adp}}$ is trained using the contrastive objective from~\cref{eq:infonce}, while the new embedding model $\Memb{}'$ and both \Mprompt{} and the original projection $\mathbf{W}_{\text{proj}}$ remain fixed.

\begin{table*}[t]
\centering
\resizebox{\textwidth}{!}{%
\begin{tabular}{l cccccccc| cccccccc}
\toprule
& \multicolumn{8}{c}{\textit{\textbf{Backbone: Llama-3.2-1B}}} 
& \multicolumn{8}{c}{\textit{\textbf{Backbone: Llama-3.2-3B}}} \\
\cmidrule(lr){2-9} \cmidrule(lr){10-17}
\textbf{Method} & Retr. & Rerank. & Clust. & PairClass. & Class. & STS & Summ. & Avg. 
                & Retr. & Rerank. & Clust. & PairClass. & Class. & STS & Summ. & Avg. \\
\midrule
\multicolumn{17}{l}{\textit{\textcolor{gray}{Training-free Discrete Prompting}}} \\
Vanilla
  & 8.96 & 25.07 & 41.00 & 47.36 & 50.67 & 50.38 & -2.11 & 35.38
  & 9.24 & 25.22 & 41.67 & 50.76 & 51.57 & 50.60 & -3.27 & 36.02  \\
PromptEOL
  & 17.56 & 28.24 & 43.00 & 51.62 & 68.57 & 62.55 & 8.37  & 44.88
  & 19.27 & 31.16 & 47.94 & 67.24 & 73.24 & 67.30 & 22.64 & 49.82 \\
Pretend-CoT
  & 14.24 & 27.70 & 42.04 & 50.68 & \textbf{71.23} & 62.35 & 11.97 & 44.30
  & 19.33 & 31.81 & 45.75 & 57.62 & 72.32 & 70.62 & 19.05 & 49.18 \\
Echo
  & 14.21 & 29.36 & 44.26 & \textbf{63.20} & 62.05 & 61.93 & 8.53  & 43.70
  & 16.31 & 29.58 & 44.83 & 67.48 & 63.92 & 65.06 & 8.51  & 45.74 \\
\midrule
\multicolumn{17}{l}{\textit{\textcolor{gray}{Soft-Prompting Methods}}} \\
Prompt-Tuning
  & 66.97 & 39.66 & 76.03 & 41.37 & 28.49 & 68.17 & 24.89 & 50.90
  & 42.18 & 45.61 & 51.29 & 76.21 & 72.75 & \textbf{78.03} & 29.90 & 60.15 \\
\rowcolor{ourblue}
\model{} 
  & \textbf{76.37} & \textbf{50.72} & \textbf{77.40} & 47.42 & 41.59 & \textbf{76.45} & \textbf{32.16} & \textbf{61.18}
  & \textbf{46.87} & \textbf{48.58} & \textbf{53.42} & \textbf{81.30} & \textbf{80.39} & 77.70 & \textbf{35.98} & \textbf{63.79} \\
\midrule
\multicolumn{17}{l}{\textit{\textcolor{gray}{Fine-tuning}}} \\
\rowcolor{loragray}
LoRA
  & 81.96 & 52.76 & 82.57 & 47.71 & 48.13 & 75.48 & 31.49 & 63.73
  & 48.57 & 47.23 & 51.69 & 84.18 & 80.47 & 77.19 & 34.63 & 63.89 \\
\midrule\midrule
& \multicolumn{8}{c}{\textit{\textbf{Backbone: Qwen2.5-7B}}} 
& \multicolumn{8}{c}{\textit{\textbf{Backbone: Mistral-7B}}} \\
\cmidrule(lr){2-9} \cmidrule(lr){10-17}
\textbf{Method} & Retr. & Rerank. & Clust. & PairClass. & Class. & STS & Summ. & Avg. 
                & Retr. & Rerank. & Clust. & PairClass. & Class. & STS & Summ. & Avg. \\
\midrule
\multicolumn{17}{l}{\textit{\textcolor{gray}{Training-free Discrete Prompting}}} \\
Vanilla
  & 7.40 & 28.23 & 40.98 & 49.75 & 43.59 & 45.02 & -2.12 & 32.65
  & 8.30 & 25.85 & 40.16 & 50.24 & 51.37 & 48.67 & -2.93 & 35.01 \\
PromptEOL
  & 24.48 & 32.78 & 46.13 & 71.36 & 69.65 & 67.00 & 6.22 & 50.04
  & 21.64 & 30.58 & 43.52 & 61.95 & 72.68 & 67.01 & 15.75 & 48.86 \\
Pretend-CoT
  & 26.06 & 31.15 & 47.23 & 68.71 & 74.76 & 68.98 & 23.02 & 52.29
  & 21.12 & 30.19 & 43.14 & 67.09 & 75.49 & 69.07 & 23.33 & 50.19 \\
Echo
  & 10.29 & 30.58 & 39.38 & 66.74 & 55.24 & 57.22 & 6.40 & 39.57
  & 13.82 & 29.83 & 42.89 & 66.56 & 62.79 & 61.74 & 4.97  & 43.60 \\
\midrule
\multicolumn{17}{l}{\textit{\textcolor{gray}{Soft-Prompting Methods}}} \\
Prompt-Tuning
  & 39.37 & 46.95 & 50.71 & 76.47 & 70.74 & 75.19 & 28.04 & 58.38
  & 47.38 & 48.53 & 53.96 & 80.59 & 78.10 & 78.22 & \textbf{36.01} & 63.64 \\
\rowcolor{ourblue}
\model{} 
  & \textbf{47.88} & \textbf{49.73} & \textbf{55.03} & \textbf{83.15} & \textbf{79.88} & \textbf{75.70} & \textbf{31.32} & \textbf{63.89}
  & \textbf{50.42} & \textbf{49.77} & \textbf{55.16} & \textbf{84.11} & \textbf{79.75} & \textbf{78.26} & 34.77 & \textbf{65.23} \\
\midrule
\multicolumn{17}{l}{\textit{\textcolor{gray}{Fine-tuning}}} \\
\rowcolor{loragray}
LoRA
  & 51.10 & 47.49 & 55.26 & 84.46 & 80.10 & 74.71 & 33.21 & 64.18
  & 55.24 & 49.21 & 54.28 & 85.65 & 84.36 & 73.98 & 36.31 & 66.32 \\
\bottomrule
\end{tabular}%
}
\caption{
  Main results on the MTEB (English, v2) benchmark across four representative LLM backbones. Methods are grouped by their adaptation strategy: training-free discrete prompting, soft-prompting, and parameter-efficient fine-tuning. \textbf{Bold} highlights the best performance among frozen-backbone methods.
}
\label{tab:embedding-benchmark}
\vspace{-10pt}
\end{table*}
\section{Experiment}

\subsection{Experimental Setup}

\noindent\textbf{Backbone LLMs.}
We evaluate \model{} on a range of pretrained LLMs spanning 1B to 8B parameters: Llama-3.2-1B, Llama-3.2-3B, Llama-3.1-8B, Mistral-7B, and Qwen-2.5-7B.

\noindent\textbf{Evaluation.}
We evaluate on MTEB (English, v2)~\cite{MTEBv2}, a comprehensive benchmark covering 41 datasets across seven task categories. For a more extensive comparison, we have also evaluated on instruction following benchmark~\cite{feng2025don}. Detailed results and the specific instructions used for each dataset are provided in Appendix~\ref{appendix:instruction-following} and Appendix~\ref{appendix:full-instruct}, respectively.

\noindent\textbf{Baselines.}
We compare \model{} against three categories of methods.
\textbf{(1) Training-free prompting:}  
comprising Vanilla (direct inference without any instructional template) alongside PromptEOL~\cite{prompteol}, Pretend-CoT~\cite{pretentCOT}, and Echo~\cite{echo}. These methods elicit embeddings from LLMs via carefully designed prompts without any parameter updates.
\textbf{(2) Prompt tuning}~\cite{prompt_tuning}: The soft-prompt baseline that learns a fixed set of continuous prefix vectors prepended to the input, with all LLM parameters frozen.
\footnote{We also attempted to train layer-wise soft prompts like P-Tuning v2~\cite{ptuningv2} as an additional baseline; however, we found it consistently failed to converge and yield degenerate embeddings. We therefore exclude it from our main comparison.}
\textbf{(3) LoRA fine-tuning:} Parameter-efficient adaptation via low-rank weight updates~\cite{lora} applied directly to the pretrained LLM.
\textbf{(4) Off-the-shelf embedding models:} including E5-large~\cite{E5-large}, GTE-large~\cite{gte}, LLM2Vec~\cite{LLM2vec} and GRITLM~\cite{GRIT}.

\noindent\textbf{Training Details.}
We utilize the dataset from \cite{bge-icl}, which contains supervised pairs and hard negatives for contrastive learning. To balance training efficiency and computational costs, we sample 10\% (0.2M) of the original data for our experiments following prior work~\cite{gircse}. 
We use Qwen3-0.6B as our default prompting model and set the number of soft prompting tokens to $k=5$. The implementation details and hyperparameters for \model{} and baseline methods are provided in Appendix~\ref{appendix:implement_detail}.

\subsection{Main Results}
We first compare the embedding performance of \model{} with other baseline methods on the MTEB (English, v2) benchmark. The results are presented in Table~\ref{tab:embedding-benchmark}. We highlight two key observations.

\noindent\textbf{\model{} consistently achieves the best performance among frozen-backbone methods.}
Although training-free prompting methods such as PromptEOL and Echo require no training data and could be applied to arbitrary pretrained LLMs, they suffer a substantial performance gap compared to soft-prompting methods (e.g., PromptEOL at 44.88 vs.\ \model{} at 61.18 on Llama-3.2-1B). 
This demonstrates the necessity of additional training data and the importance of prompt tuning. Among soft-prompting methods, \model{} outperforms Prompt-Tuning across all four backbones by a consistent and significant margin (e.g., +10.3 on Llama-3.2-1B, +5.5 on Qwen2.5-7B, +3.6 on Llama-3.2-3B). We attribute this advantage to the aid of the prompting LLM, which generates dynamic and semantically rich tokens that more effectively steer the frozen backbone for generating effective embeddings.

\noindent \textbf{\model{} achieves competitive performance to LoRA fine-tuning without modifying the backbone.}
Without any parameter updates to the backbone LLM, \model{} remains within 2.6 average scores of LoRA across all backbones. On Llama-3.2-3B and Qwen2.5-7B, the gap narrows to just 0.1 and 0.3 average scores, respectively. This demonstrates that inserting only a few soft tokens is sufficient to achieve performance comparable to weight-level adaptation while fully preserving the generality of the backbone.

\subsection{Comparison with Off-the-shelf Embedding Models}
We further compare \model{} among representative off-the-shelf embedding models on MTEB, including encoder-based models (E5-Large, GTE-Large), and LLM-based methods (LLM2Vec, E5-Mistral, GRITLM). As illustrated in Table~\ref{tab:off_the_shelf_models}, \model{} using Mistral-7B backbone outperforms all encoder-based baselines and LLM2Vec, achieving a superior average score of 65.2. Remarkably, even without any parameter updates to the backbone, \model{} remains on par with state-of-the-art models like E5-Mistral (66.3) and GRITLM (66.9). This performance is particularly noteworthy as it demonstrates that our frozen-backbone paradigm can rival heavily fine-tuned models.

\begin{table}[t]
\centering
\resizebox{\columnwidth}{!}{%
\begin{tabular}{l ccccccc|c}
\toprule
\textbf{Method} & Retr. & Rerank. & Clust. & PairClass. & Class. & STS & Summ. & Avg. \\
\midrule
E5-Large             & 49.3 & 45.7 & 45.2 & 86.1 & 76.4 & 80.7 & 32.3 & 62.8  \\
GTE-Large            & 53.3 & 47.8 & 48.2 & 85.1 & 75.5 & 83.3 & 32.9 & 64.8  \\
LLM2Vec              & 51.3 & 47.7 & 44.1 & 88.0 & 79.7 & 83.7 & 31.1 & 64.6  \\
E5-Mistral$^\dagger$ & 55.2 & 49.2 & 54.3 & 85.7 & 84.4 & 74.0 & 36.3 & 66.3  \\
GRITLM$^\dagger$     & 55.4  & 48.7 & 54.6 & 86.3 & 84.9 & 75.9 & 36.1 & 66.9  \\
\midrule
\rowcolor{ourblue}
\model{} 
  & 50.4 & 49.8 & 55.2 & 84.1 & 79.8 & 78.3 & 34.8 & 65.2 \\
\bottomrule
\end{tabular}%
}
\caption{Comparison of \model{} with Mistral-7B backbone against off-the-shelf embedding models on MTEB. 
$^\dagger$ indicates reproduced results with the same training data to ensure a fair baseline comparison.
}
\label{tab:off_the_shelf_models}
\vspace{-15pt}
\end{table}

\begin{table*}[t]
\centering
\resizebox{0.93\linewidth}{!}{
\begin{tabular}{ll *{4}{S[table-format=2.2]} S[table-format=2.2] S[table-format=2.2]}
\toprule
& & \multicolumn{4}{c}{\textbf{Target LLM}} & & \\
\cmidrule(lr){3-6}
\textbf{Source LLM} & \textbf{Method} 
  & {\textbf{Llama-3.2 (3B)}} & {\textbf{Llama-3.1 (8B)}} & {\textbf{Mistral (7B)}} & {\textbf{Qwen2.5 (7B)}}
  & {\textbf{Avg.}} & {\textbf{$\Delta$ vs.\ LoRA}} \\
\midrule

\rowcolor[gray]{0.93} \multicolumn{8}{l}{\textit{Training-free Prompting}} \\
{---} & PromptEOL & 49.11 & 49.40 & 48.86 & 50.04 & 49.35 & {$-$16.21} \\
{---} & Echo      & 45.74 & 44.42 & 43.60 & 39.57 & 43.26 & {$-$22.30} \\
\midrule

\rowcolor[gray]{0.93} \multicolumn{8}{l}{\textit{In-model Fine-tuning}} \\
{---} & LoRA      & 65.19 & 66.56 & 66.32 & 64.18 & 65.56 & {---} \\
\midrule

\rowcolor[gray]{0.93} \multicolumn{8}{l}{\textit{Cross-model Transfer}} \\
\multirow{2}{*}{Llama-3.1 (8B)} 
  & Prompt-Tuning & 60.79          & \underline{63.38} & 64.26          & 57.93          & 61.59 & {$-$3.97} \\
  & \cellcolor{ourblue}\model{} & \cellcolor{ourblue}\textbf{63.16} & \cellcolor{ourblue}\underline{\textbf{64.86}} & \cellcolor{ourblue}\textbf{64.87} & \cellcolor{ourblue}\textbf{63.97} & \cellcolor{ourblue}\textbf{64.22} & \cellcolor{ourblue}\textbf{$-$1.34} \\
\addlinespace[3pt]

\multirow{2}{*}{Mistral (7B)}   
  & Prompt-Tuning & 60.68          & 60.49             & \underline{63.64} & 51.72          & 59.13 & {$-$6.43} \\
  & \cellcolor{ourblue}\model{} & \cellcolor{ourblue}\textbf{62.92} & \cellcolor{ourblue}\textbf{64.19}    & \cellcolor{ourblue}\underline{\textbf{65.22}} & \cellcolor{ourblue}\textbf{63.11} & \cellcolor{ourblue}\textbf{63.86} & \cellcolor{ourblue}\textbf{$-$1.70} \\
\addlinespace[3pt]

\multirow{2}{*}{Qwen2.5 (7B)}   
  & Prompt-Tuning & 59.71          & 62.10             & \textbf{64.90} & \underline{58.38} & 61.27 & {$-$4.29} \\
  & \cellcolor{ourblue}\model{} & \cellcolor{ourblue}\textbf{63.08} & \cellcolor{ourblue}\textbf{64.08}    & \cellcolor{ourblue}64.14          & \cellcolor{ourblue}\underline{\textbf{63.89}} & \cellcolor{ourblue}\textbf{63.80} & \cellcolor{ourblue}\textbf{$-$1.76} \\
\bottomrule
\end{tabular}
}
\caption{
  Cross-model transferability on MTEB (English, v2).
  \underline{Underlined} values represent the non-transfer cases (i.e., train from scratch).
  \textbf{Bold} indicates the superior performance among the cross-model transfer methods.
  \textbf{Avg.}\ is computed over all four target LLMs for fairness.
  $\Delta$\textbf{ vs.\ LoRA} measures the gap to the LoRA fine-tuning average.
}
\label{tab:transferability_results_full}
\vspace{-10pt}
\end{table*}
\section{Embedding Model Transferability}

\subsection{Cross-Architecture Transferability}
\label{subsec:cross_model_transfer}
We evaluate cross-architecture transferability by comparing \model{} transfer against Prompt-Tuning transfer across different source to target backbone settings. For reference, we also report LoRA fine-tuning performance trained from scratch on each target backbone. Results are presented in Table~\ref{tab:transferability_results_full}, where we highlight three key observations.
\noindent\textbf{1) \model{} transfers substantially better than Prompt-Tuning.}
Across all transfer settings, \model{} consistently outperforms Prompt-Tuning transfer by a large margin. The average gap to LoRA is reduced from 3.97 to 6.43 average scores, down to just 1.34 to 1.76 average scores. This demonstrates that the auxiliary prompting LLM learns more transferable representations than conventional soft prompts.
\noindent\textbf{2) Transfer performance closely matches training from scratch.}
The performance drop when transferring to a different model backbone is minimal. When using Mistral-7B as the target, \model{} achieves 65.22 when trained from scratch, and still maintains 64.87 and 64.14 when transferred from Llama-3.1 and Qwen2.5, respectively. Notably, \model{}'s average performance remains within 1.5\% of LoRA, which requires fine-tuning on the target model from scratch.  This shows that \model{} learns generalizable representations that are agnostic to the source LLM used during initial training.

\begin{figure}[t]
    \centering
    \includegraphics[width=\linewidth]{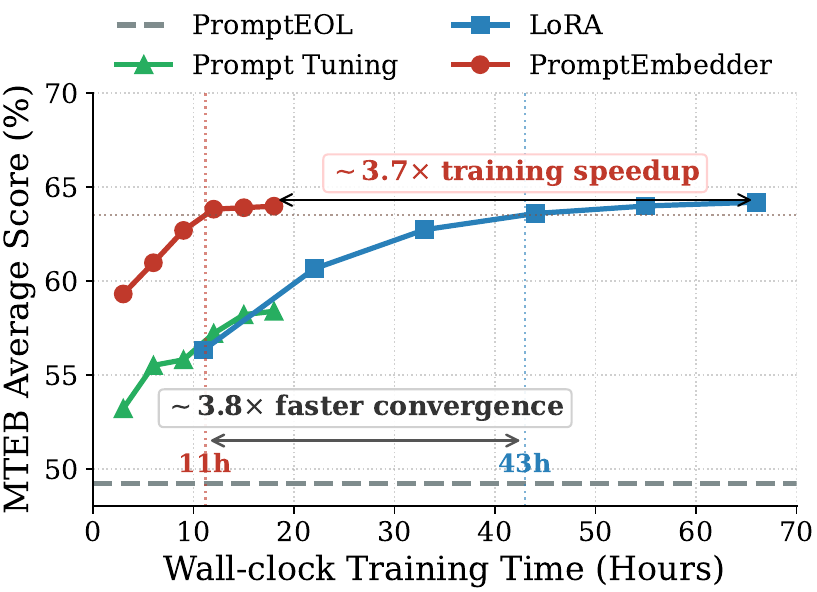}
    \caption{
    Training efficiency comparison: MTEB performance vs. wall-clock time. \model{} achieves a \textbf{3.7$\times$ speedup} over LoRA in reaching comparable performance levels.
    }
    \label{fig:learning-eff}
    \vspace{-25pt}
\end{figure}

\subsection{Transfer Efficiency on Training Cost}
\label{subsec:training_efficiency}

Beyond superior transfer performance, \model{} offers a significant advantage in computational efficiency during cross-model transfer. To evaluate this, we perform a cross-model transfer from a Llama3-8B source model to a Qwen2.5-7B target model.  We compare the wall-clock training time of \model{} against LoRA and Prompt-Tuning by tracking MTEB performance over training hours.\footnote{To ensure a fair and reproducible comparison, all experiments are conducted on a single NVIDIA RTX 3090 GPU.} 

As illustrated in Figure~\ref{fig:learning-eff}, \model{} demonstrates two efficiency gains. First, \model{} achieves \textbf{3.8$\times$ faster convergence} compared to LoRA. Specifically, \model{} reaches an MTEB score of 63.81 after only 11 hours of training and plateaus shortly thereafter, whereas LoRA requires 43 hours to achieve a comparable performance level of 63.58. We attribute this accelerated convergence to the embedding knowledge stored within \Mprompt{}, which can be effectively transferred to arbitrary LLM backbones without extensive retraining.
Second, \model{} delivers a \textbf{3.7$\times$ training time speedup} per epoch compared to LoRA. In our experiments, \model{} requires only 18 hours to iterate over the entire training dataset, while LoRA requires 66 hours to complete the same process. This speedup stems from the lightweight fine-tuning of the adaptation matrix $\mathbf{W}_{\text{adp}}$. These results demonstrate that \model{} significantly reduces the computational overhead of cross-model transfer while maintaining high performance.

\section{Discussion}

\begin{figure}[t]
    \centering
    \includegraphics[width=0.85\linewidth]{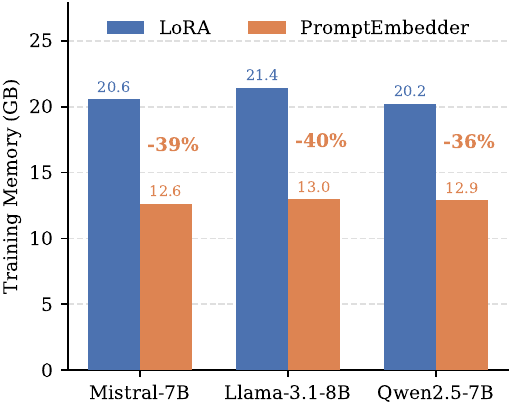}
    \vspace{-5pt}
    \caption{Comparison of peak GPU memory usage between LoRA and \model{} across three representative LLM backbones.}
    \label{fig:training-mem}
    \vspace{-15pt}
\end{figure}

\subsection{Memory Efficiency}
We evaluate the peak GPU memory consumption of \model{} during training and compare it against LoRA across three representative LLM backbones. As illustrated in Figure~\ref{fig:training-mem}, \model{} reduces training memory by 36\% to 40\% across all backbones. This significant reduction is primarily attributed to the fact that no gradients are propagated through the pretrained LLM \Memb{}. Combined with the training time efficiency demonstrated in Section~\ref{subsec:training_efficiency}, these results validate \model{} as a resource-efficient alternative to LoRA fine-tuning.

\subsection{Number of Prompting Tokens}
 We investigate how the number of soft tokens affects the embedding quality of both \model{} and Prompt-Tuning. We varied the number of tokens $k \in \{1,3,5,10,15\}$ and summarized the results in Figure~\ref{fig:token-performance}. In general, performance improves consistently as the number of soft tokens increases for both methods. However, \model{} consistently outperforms Prompt-Tuning across different numbers of soft tokens. For instance, with only 5 soft tokens, \model{} achieves an MTEB score of 64.86, compared to 57.40 for Prompt-Tuning. 
 Furthermore, \model{} achieves near-optimal performance at $k{=}5$, with negligible gains beyond this point. Based on these results, we set $k{=}5$ as the default configuration in all experiments.

\begin{figure}[t]
\centering
\begin{tikzpicture}
\begin{axis}[
    width=\linewidth,
    height=0.55\linewidth,
    xlabel={Number of Tokens},
    ylabel={MTEB Score},
    grid=major,
    grid style={dashed, line width=0.2pt, draw=gray!40},
    axis line style={draw=black!80},
    tick style={draw=black!80},
    tick align=outside,
    symbolic x coords={1,3,5,10,15},
    xtick=data,
    ymin=50, ymax=67,
    ytick={50, 55, 60, 65, 70},
    label style={font=\small},
    tick label style={font=\small},
    legend style={
        at={(0.5,1.08)},
        anchor=south,
        legend columns=3,
        draw=none,
        fill=none,
        font=\small,
        /tikz/every even column/.append style={column sep=5pt}
    },
]

\definecolor{PromptRed}{HTML}{A63426} 
\definecolor{LoRABlue}{HTML}{4D51B0}  
\definecolor{BaselineGray}{HTML}{999999}

\addplot[
    color=BaselineGray,
    dashed,
    line width=1pt,
    forget plot
] coordinates {(1, 49.2) (15, 49.2)}; 

\addplot[
    color=PromptRed,
    very thick,
    mark=*,
    mark size=2.4pt,
] coordinates {
    (1, 60.37)
    (3, 63.37)
    (5, 64.86)
    (10, 64.88)
    (15, 64.79)
};
\addlegendentry{\model{}}

\addplot[
    color=LoRABlue,
    very thick,
    mark=square*,
    mark size=2.4pt,
] coordinates {
    (1, 52.90)
    (3, 53.90)
    (5, 57.33)
    (10, 59.51)
    (15, 60.10)
};
\addlegendentry{Prompt-Tuning}

\end{axis}
\end{tikzpicture}
\caption{MTEB performance by varying the number of soft prompting tokens with Llama-8B backbone.}
\label{fig:token-performance}
\end{figure}
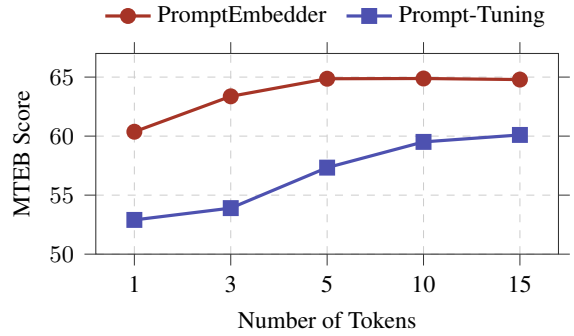
\begin{table}[t]
\centering
\resizebox{\columnwidth}{!}{%
\begin{tabular}{lcccc}
\toprule
 & \multicolumn{4}{c}{\textbf{Prompting LLM \Mprompt{}}} \\
\cmidrule(lr){2-5}
\textbf{Pretrained LLM \Memb{}} & \textbf{0.6B} & \textbf{1.7B} & \textbf{4B} & \textbf{8B} \\
\midrule
Mistral-7B  & 65.23 & 65.48 & 65.31 & 65.72 \\
Llama-8B  & 64.86 & 64.92 & 64.66 & 64.71 \\
\bottomrule
\end{tabular}%
}
\caption{MTEB scores (\%) across different combinations of Prompting LLM and pretrained LLM. The Prompting LLM size is varied from 0.6B to 8B using the Qwen3 family~\cite{yang2025qwen3}.}
\label{tab:prompting_llm_choice}
\vspace{-15pt}
\end{table}
\subsection{Choice of Prompting LLM}
To investigate the sensitivity of \model{} to the capacity of the prompting LLM \Mprompt{}, we evaluate the performance of \model{} using different scales of pretrained LLMs from the Qwen3 family, spanning from 0.6B to 8B parameters. As shown in Table~\ref{tab:prompting_llm_choice}, we observe remarkable performance stability across different scales of \Mprompt{}. For instance, scaling the prompting model from 0.6B to 8B yields only a marginal improvement of 0.49\% for the Mistral-7B backbone. This finding demonstrates that the generation of effective soft prompts does not necessitate a large-scale LLM. Consequently, lightweight models (e.g., 0.6B) can be utilized to maintain high efficiency without compromising the quality of embeddings.

\section{Conclusion}
We introduce \model{}, a dual-LLM framework that separates embedding-specific knowledge from the pretrained LLMs. By using an auxiliary Prompting LLM to generate differentiable soft prompts, \model{} enables high-quality representation learning with a frozen pretrained LLM and efficient transferability. Extensive experiments show that \model{} outperforms existing prompting-based methods and achieves performance comparable to LoRA-based fine-tuning while significantly improving training efficiency. We believe \model{} offers a practical and scalable paradigm for LLM-based text embeddings.
\newpage

\section{Limitations}
While \model{} substantially closes the gap to LoRA fine-tuning, a small performance difference of up to 2.6 average score remains on certain backbones. We note that this gap is a direct consequence of keeping \Memb{} entirely frozen, a deliberate design choice that enables cross-architecture transferability and significant reductions in memory and training cost. Additionally, the current cross-architecture transfer setting requires a small amount of labeled data to train the linear alignment layer. Exploring unsupervised or zero-shot transfer to unseen backbones is a promising direction for future work.


\bibliography{custom}

\appendix
\appendix



\section{Use-of-LLMs}

In this work, we utilized large language models (LLMs) as part of the core research methodology. Specifically, we fine-tuned existing open-source LLMs (e.g., LLaMA-3 and Mistral) to develop embedding models. These pre-trained models served as the foundation for our experiments, and our main contributions build upon their architectures and representations. Additionally, a LLM-based assistant was used for minor writing support, including grammar checking and improving manuscript readability. All decisions regarding research design, fine-tuning strategies, experimental setup, and final interpretations were made solely by the authors.

\section{Implementation Details}
\label{appendix:implement_detail}
For both \model{} and the baseline models we re-implement for comparison, we use nearly identical fine-tuning hyperparameters across different model sizes.
We use the AdamW optimizer with a learning rate of 1e-4 and a warmup ratio of 0.03. We adopt Low-Rank Adaptation (LoRA)~\cite{lora} for efficient fine-tuning, setting the rank to 64 and the scaling factor $\alpha$ to 16. Due to limited computational resources, we train with a batch size of 2 and accumulate gradients over 8 steps, resulting in an effective batch size of 16. All models are fine-tuned for a single epoch. For prompt-tuning baseline, we optimize 20 virtual tokens throughout the experiment.

\section{Evaluation on Instruction Following Benchmark}
\label{appendix:instruction-following}
Table~\ref{tab:appendix_full_instruction_following} provides the instruction following performance breakdown across STS, Triplet Alignment, and Clustering tasks. The results demonstrate that \model{} consistently outperforms all frozen-backbone baselines, including training-free prompting and Prompt-Tuning, across all four evaluated architectures. For instance, on the LLaMA-3.2-3B backbone, \model{} achieves an average score of 44.37, representing a significant improvement over the prompt-tuning method with a score of 36.71. Most notably, \model{} exhibits high competitiveness even when compared against LoRA. Using LLaMA-3.2-1B as backbone, \model{} reaches an average of 44.07, actually surpassing LoRA (43.65). For larger-scale models such as Qwen2.5-7B, \model{} remains remarkably robust with a score of 42.20, narrowing the performance gap with LoRA (43.12) to within 1 point.

\begin{table*}[t]
\centering
\resizebox{\textwidth}{!}{%
\begin{tabular}{l ccc ccc cc | ccc ccc cc}
\toprule
& \multicolumn{8}{c}{\textit{\textbf{Backbone: LLaMA-3.2-1B}}} & \multicolumn{8}{c}{\textit{\textbf{Backbone: LLaMA-3.2-3B}}} \\
\cmidrule(lr){2-9} \cmidrule(lr){10-17}
& \multicolumn{3}{c}{STS $\uparrow$} & \multicolumn{3}{c}{Triplet Align. $\uparrow$} & \multicolumn{1}{c}{Clust. $\uparrow$} & & \multicolumn{3}{c}{STS $\uparrow$} & \multicolumn{3}{c}{Triplet Align. $\uparrow$} & \multicolumn{1}{c}{Clust. $\uparrow$} & \\
\cmidrule(lr){2-4} \cmidrule(lr){5-7} \cmidrule(lr){8-8} \cmidrule(lr){10-12} \cmidrule(lr){13-15} \cmidrule(lr){16-16}
\textbf{Method} & PC & MH & BP & IE & Tox. & AG & NYC & Avg. & PC & MH & BP & IE & Tox. & AG & NYC & Avg. \\
\midrule
\multicolumn{17}{l}{\textit{\textcolor{gray}{Training-free Prompting}}} \\
PromptEOL  & 24.61 & 9.94  & 23.65 & 15.26 & 52.29 & 68.10 & 2.77  & 25.52
           & 24.37 & 11.21 & 23.48 & 11.08 & 53.79 & 71.00 & 1.87  & 24.90 \\
Pretend-CoT & 21.41 & 10.71 & 22.27 & 10.70 & 54.15 & 66.21 & 3.21 & 24.58
            & 26.82 & 15.41 & 21.25 & 17.37 & 54.27 & 72.87 & 2.73  & 26.68 \\
Echo       & \textbf{33.13} & 6.14  & 20.74 & 20.74 & 52.68 & 76.13 & 37.18 & 37.82
           & 36.37 & 6.63  & 22.38 & 28.38 & 52.28 & 75.74 & 39.90 & 39.51 \\
\midrule
\multicolumn{17}{l}{\textit{\textcolor{gray}{Frozen Backbone Methods}}} \\
Prompt-Tuning & 29.23 & 3.93  & 20.21 & 37.05 & \textbf{54.63} & 72.80 & 4.96  & 29.64
              & \textbf{39.50} & \textbf{16.70} & 26.47 & 57.18 & 54.35 & 80.62 & 3.40  & 36.71 \\
\rowcolor{ourblue}
\model{}  & 27.74 & \textbf{13.30} & \textbf{24.85} & \textbf{53.42} & 54.59 & \textbf{83.62} & \textbf{42.12} & \textbf{44.07}
                & 28.16 & 8.86  & \textbf{26.92} & \textbf{59.15} & \textbf{59.58} & \textbf{84.17} & \textbf{39.75} & \textbf{44.37} \\
\midrule
\multicolumn{17}{l}{\textit{\textcolor{gray}{Fine-tuning}}} \\
\rowcolor{loragray}
LoRA & 30.61 & 16.76 & 24.85 & 53.56 & 60.71 & 80.37 & 37.72 & 43.65
               & 42.56 & 26.29 & 26.91 & 42.04 & 56.27 & 80.92 & 39.95 & 44.53 \\
\midrule\midrule
& \multicolumn{8}{c}{\textit{\textbf{Backbone: Qwen2.5-7B}}} & \multicolumn{8}{c}{\textit{\textbf{Backbone: Mistral-7B}}} \\
\cmidrule(lr){2-9} \cmidrule(lr){10-17}
& \multicolumn{3}{c}{STS $\uparrow$} & \multicolumn{3}{c}{Triplet Align. $\uparrow$} & \multicolumn{1}{c}{Clust. $\uparrow$} & & \multicolumn{3}{c}{STS $\uparrow$} & \multicolumn{3}{c}{Triplet Align. $\uparrow$} & \multicolumn{1}{c}{Clust. $\uparrow$} & \\
\cmidrule(lr){2-4} \cmidrule(lr){5-7} \cmidrule(lr){8-8} \cmidrule(lr){10-12} \cmidrule(lr){13-15} \cmidrule(lr){16-16}
\textbf{Method} & PC & MH & BP & IE & Tox. & AG & NYC & Avg. & PC & MH & BP & IE & Tox. & AG & NYC & Avg. \\
\midrule
\multicolumn{17}{l}{\textit{\textcolor{gray}{Training-free Prompting}}} \\
PromptEOL  & 28.92 & 11.09 & 26.23 & 16.76 & 54.94 & 78.57 & 2.21 & 31.25
           & 26.14 & 14.07 & 21.70 & 21.82 & 53.77 & 72.63 & 2.31  & 26.91 \\

Pretend-CoT & 28.47 & 14.63 & 25.67 & 17.94 & 56.07 & 76.77 & 5.14 & 32.10
            & 30.31 & 19.26 & 21.17 & 20.33 & 53.94 & 71.81 & 3.20  & 28.84 \\

Echo       & 26.10 & 5.32 & 21.74 & 14.96 & 53.37 & 78.54 & \textbf{25.94} & 32.28
           & 28.41 & 6.00  & 19.79 & 22.24 & \textbf{56.41} & 76.11 & \textbf{31.57} & 34.54 \\

\midrule
\multicolumn{17}{l}{\textit{\textcolor{gray}{Frozen Backbone Methods}}} \\
Prompt-Tuning & 33.97 & \textbf{18.26} & \textbf{27.38} & \textbf{57.59} & 52.81 & 81.91 & 5.08 & 39.57
              & \textbf{39.32} & \textbf{21.49} & \textbf{23.91} & 53.10 & 55.55 & 81.60 & 2.08  & 35.03 \\

\rowcolor{ourblue}
\model{}  & \textbf{38.44} & 15.46 & 25.90 & 46.65 & \textbf{58.95} & \textbf{85.36} & 24.66 & \textbf{42.20}
                & 21.72 & 7.40  & 18.60 & \textbf{65.01} & 53.78 & \textbf{82.24} & 23.85 & \textbf{37.87} \\

\midrule
\multicolumn{17}{l}{\textit{\textcolor{gray}{Fine-tuning}}} \\
\rowcolor{loragray}
LoRA & 39.52 & 18.84 & 26.15 & 45.30 & 58.12 & 85.90 & 28.01 & 43.12
               & 44.26 & 15.25 & 18.83 & 35.33 & 60.09 & 84.81 & 58.76 & 44.03 \\
\bottomrule
\end{tabular}%
}
\caption{
  Evaluation on instruction-following embedding tasks across four LLM backbones,
  covering Semantic Textual Similarity (STS), Triplet Alignment, and Clustering.
  \textbf{Bold} denotes the best score among frozen-backbone methods.
  Dataset abbreviations: PC (PaperCode), MH (MultiHate), BP (Big Patent),
  IE (IntentEmo), Tox.\ (Toxic), AG (AG-News), NYC (NYTClust).
}
\label{tab:appendix_full_instruction_following}
\end{table*}
\section{Full Dataset Instructions}
\label{appendix:full-instruct}
This section provides the complete set of natural language instructions used with the datasets in our experiments. We include the instructions for the MTEB benchmark datasets. These instructions define the intended task for each dataset and serve as the input prompts during embedding evaluation. The full text of the instructions is listed in Table~\ref{tab:mteb_instruct}.

This section provides the complete set of natural language instructions used with the datasets in our experiments. We include the instructions for both the MTEB benchmark datasets and instruction-following embedding tasks. These instructions define the intended task or target aspect for each dataset and serve as the input prompts during embedding evaluation. The full text of the instructions is listed in Table~\ref{tab:mteb_instruct} and Table~\ref{tab:instruction_following_instruct}.

\onecolumn
\begingroup
\small

\begin{longtable}{p{0.34\linewidth} p{0.60\linewidth}}
\caption{Instructions for the corresponding datasets in the MTEB benchmark. We mainly follow the instructions from the \textbf{GritLM} paper. Note that for retrieval and reranking datasets, queries (Q) and corpus (C) documents may require different instructions, denoted as \{dataset\}-\textbf{Q} and \{dataset\}-\textbf{C}, respectively. For datasets with query instructions only (i.e., \{dataset\}-\textbf{Q}), no instructions are applied to the corpus.}\\
\hline
\textbf{Dataset} & \textbf{Instruction} \\
\hline
\endfirsthead
\multicolumn{2}{c}%
{{\bfseries \tablename\ \thetable{} -- continued from previous page}} \\
\hline
\textbf{Dataset} & \textbf{Instruction} \\
\hline
\endhead
\hline \multicolumn{2}{r}{{Continued on next page}} \\
\endfoot
\hline
\endlastfoot

SummEvalSummarization & Given a news summary, retrieve other semantically similar summaries. \\
ArXivHierarchicalClusteringP2P & Identify the main and secondary category of Arxiv papers based on the titles and abstracts. \\
ArXivHierarchicalClusteringS2S & Identify the main and secondary category of Arxiv papers based on the titles. \\
Touche2020Retrieval.v3-\textbf{Q} & Given a question, retrieve passages that answer the question. \\
ClimateFEVERHardNegatives-\textbf{Q} & Given a claim about climate change, retrieve documents that support or refute the claim. \\
FEVERHardNegatives-\textbf{Q} & Given a claim, retrieve documents that support or refute the claim. \\
HotpotQAHardNegatives-\textbf{Q} & Given a multi-hop question, retrieve documents that can help answer the question. \\
AmazonCounterfactualClassification & Classify a given Amazon customer review text as either counterfactual or not-counterfactual. \\
AmazonPolarityClassification & Classify Amazon reviews into positive or negative sentiment. \\
AmazonReviewsClassification & Classify the given Amazon review into its appropriate rating category. \\
Banking77Classification & Given a online banking query, find the corresponding intents. \\
EmotionClassification & Classify the emotion expressed in the given Twitter message into one of the six emotions: anger, fear, joy, love, sadness, and surprise. \\
ImdbClassification & Classify the sentiment expressed in the given movie review text from the IMDB dataset. \\
MassiveIntentClassification & Given a user utterance as query, find the user intents. \\
MassiveScenarioClassification & Given a user utterance as query, find the user scenarios. \\
MTOPDomainClassification & Classify the intent domain of the given utterance in task-oriented conversation. \\
MTOPIntentClassification & Classify the intent of the given utterance in task-oriented conversation. \\
ToxicConversationsClassification & Classify the given comments as either toxic or not toxic. \\
TweetSentimentExtractionClassification & Classify the sentiment of a given tweet as either positive, negative, or neutral. \\
ArxivClusteringP2P & Identify the main and secondary category of Arxiv papers based on the titles and abstracts. \\
ArxivClusteringS2S & Identify the main and secondary category of Arxiv papers based on the titles. \\
BiorxivClusteringP2P & Identify the main category of Biorxiv papers based on the titles and abstracts. \\
BiorxivClusteringS2S & Identify the main category of Biorxiv papers based on the titles. \\
MedrxivClusteringP2P & Identify the main category of Medrxiv papers based on the titles and abstracts. \\
MedrxivClusteringS2S & Identify the main category of Medrxiv papers based on the titles. \\
RedditClustering & Identify the topic or theme of Reddit posts based on the titles. \\
RedditClusteringP2P & Identify the topic or theme of Reddit posts based on the titles and posts. \\
StackExchangeClustering & Identify the topic or theme of StackExchange posts based on the titles. \\
StackExchangeClusteringP2P & Identify the topic or theme of StackExchange posts based on the given paragraphs. \\
TwentyNewsgroupsClustering & Identify the topic or theme of the given news articles. \\
SprintDuplicateQuestions & Retrieve duplicate questions from Sprint forum. \\
TwitterSemEval2015 & Retrieve tweets that are semantically similar to the given tweet. \\
TwitterURLCorpus & Retrieve tweets that are semantically similar to the given tweet. \\
AskUbuntuDupQuestions-\textbf{Q} & Retrieve duplicate questions from AskUbuntu forum. \\
AskUbuntuDupQuestions-\textbf{C} & Retrieve duplicate questions from AskUbuntu forum. \\
MindSmallReranking-\textbf{Q} & Retrieve relevant news articles based on user browsing history. \\
MindSmallReranking-\textbf{C} & Retrieve relevant news articles based on user browsing history. \\
SciDocsRR-\textbf{Q} & Given a title of a scientific paper, retrieve the titles of other relevant papers. \\
SciDocsRR-\textbf{C} & Given a title of a scientific paper, retrieve the titles of other relevant papers. \\
StackOverflowDupQuestions-\textbf{Q} & Retrieve duplicate questions from StackOverflow forum. \\
StackOverflowDupQuestions-\textbf{C} & Retrieve duplicate questions from StackOverflow forum. \\
ArguAna-\textbf{Q} & Given a claim, find documents that refute the claim. \\
ClimateFEVER-\textbf{Q} & Given a claim about climate change, retrieve documents that support or refute the claim. \\
CQADupstackRetrieval-\textbf{Q} & Given a question, retrieve detailed question descriptions from Stackexchange that are duplicates to the given question. \\
DBPedia-\textbf{Q} & Given a query, retrieve relevant entity descriptions from DBPedia. \\
FEVER-\textbf{Q} & Given a claim, retrieve documents that support or refute the claim. \\
FiQA2018-\textbf{Q} & Given a financial question, retrieve user replies that best answer the question. \\
HotpotQA-\textbf{Q} & Given a multi-hop question, retrieve documents that can help answer the question. \\
MSMARCO-\textbf{Q} & Given a web search query, retrieve relevant passages that answer the query. \\
NFCorpus-\textbf{Q} & Given a question, retrieve relevant documents that best answer the question. \\
NQ-\textbf{Q} & Given a question, retrieve Wikipedia passages that answer the question. \\
QuoraRetrieval-\textbf{Q} & Given a question, retrieve questions that are semantically equivalent to the given question. \\
SCIDOCS-\textbf{Q} & Given a scientific paper title, retrieve paper abstracts that are cited by the given paper. \\
SciFact-\textbf{Q} & Given a scientific claim, retrieve documents that support or refute the claim. \\
Touche2020-\textbf{Q} & Given a question, retrieve detailed and persuasive arguments that answer the question. \\
TRECCOVID-\textbf{Q} & Given a query on COVID-19, retrieve documents that answer the query. \\
STS12 & Retrieve semantically similar text. \\
STS13 & Retrieve semantically similar text. \\
STS14 & Retrieve semantically similar text. \\
STS15 & Retrieve semantically similar text. \\
STS16 & Retrieve semantically similar text. \\
STS17 & Retrieve semantically similar text. \\
STS22 & Retrieve semantically similar text. \\
BIOSSES & Retrieve semantically similar text. \\
SICK-R & Retrieve semantically similar text. \\
STSBenchmark & Retrieve semantically similar text. \\
SummEval & Given a news summary, retrieve other semantically similar summaries. \\
CQADupstackTexRetrieval-\textbf{Q} & Represent the title of a user question to find a duplicate user question title with body from the Tex StackExchange forum. \\
CQADupstackTexRetrieval-\textbf{C} & Represent the question title with body posted by a user to find a duplicate user question title from the Tex StackExchange forum. \\
CQADupstackWebmastersRetrieval-\textbf{Q} & Represent the title of a user question to find a duplicate user question title with body from the Webmasters StackExchange forum. \\
CQADupstackWebmastersRetrieval-\textbf{C} & Represent the question title with body posted by a user to find a duplicate user question title from the Webmasters StackExchange forum. \\
CQADupstackEnglishRetrieval-\textbf{Q} & Represent the title of a user question to find a duplicate user question title with body from the English StackExchange forum. \\
CQADupstackEnglishRetrieval-\textbf{C} & Represent the question title with body posted by a user to find a duplicate user question title from the English StackExchange forum. \\
CQADupstackGamingRetrieval-\textbf{Q} & Represent the title of a user question to find a duplicate user question title with body from the Gaming StackExchange forum. \\
CQADupstackGamingRetrieval-\textbf{C} & Represent the question title with body posted by a user to find a duplicate user question title from the Gaming StackExchange forum. \\
CQADupstackGisRetrieval-\textbf{Q} & Represent the title of a user question to find a duplicate user question title with body from the Gis StackExchange forum. \\
CQADupstackGisRetrieval-\textbf{C} & Represent the question title with body posted by a user to find a duplicate user question title from the Gis StackExchange forum. \\
CQADupstackUnixRetrieval-\textbf{Q} & Represent the title of a user question to find a duplicate user question title with body from the Unix StackExchange forum. \\
CQADupstackUnixRetrieval-\textbf{C} & Represent the question title with body posted by a user to find a duplicate user question title from the Unix StackExchange forum. \\
CQADupstackMathematicaRetrieval-\textbf{Q} & Represent the title of a user question to find a duplicate user question title with body from the Mathematica StackExchange forum. \\
CQADupstackMathematicaRetrieval-\textbf{C} & Represent the question title with body posted by a user to find a duplicate user question title from the Mathematica StackExchange forum. \\
CQADupstackStatsRetrieval-\textbf{Q} & Represent the title of a user question to find a duplicate user question title with body from the Stats StackExchange forum. \\
CQADupstackStatsRetrieval-\textbf{C} & Represent the question title with body posted by a user to find a duplicate user question title from the Stats StackExchange forum. \\
CQADupstackPhysicsRetrieval-\textbf{Q} & Represent the title of a user question to find a duplicate user question title with body from the Physics StackExchange forum. \\
CQADupstackPhysicsRetrieval-\textbf{C} & Represent the question title with body posted by a user to find a duplicate user question title from the Physics StackExchange forum. \\
CQADupstackProgrammersRetrieval-\textbf{Q} & Represent the title of a user question to find a duplicate user question title with body from the Programmers StackExchange forum. \\
CQADupstackProgrammersRetrieval-\textbf{C} & Represent the question title with body posted by a user to find a duplicate user question title from the Programmers StackExchange forum. \\
CQADupstackAndroidRetrieval-\textbf{Q} & Represent the title of a user question to find a duplicate user question title with body from the Android StackExchange forum. \\
CQADupstackAndroidRetrieval-\textbf{C} & Represent the question title with body posted by a user to find a duplicate user question title from the Android StackExchange forum. \\
CQADupstackWordpressRetrieval-\textbf{Q} & Represent the title of a user question to find a duplicate user question title with body from the Wordpress StackExchange forum. \\
CQADupstackWordpressRetrieval-\textbf{C} & Represent the question title with body posted by a user to find a duplicate user question title from the Wordpress StackExchange forum.
\label{tab:mteb_instruct}
\end{longtable}
\endgroup

\begin{table*}[htbp]
\centering
\small
\renewcommand{\arraystretch}{1.4}
\caption{Instructions for the corresponding datasets in our instruction-following embedding tasks.}
\label{tab:instruction_following_instruct}
\begin{tabular*}{\linewidth}{@{\extracolsep{\fill}}llm{0.6\linewidth}@{}}
\toprule
\textbf{Dataset} & \textbf{Aspect} & \textbf{Instruction} \\
\midrule
\multirow{2}{*}[-0.3em]{PaperCode} 
 & Method & Represent the paper based on the method used. \\
\cmidrule{2-3}
 & Task & Represent the paper abstract based on the research task. \\
\midrule
\multirow{2}{*}[-0.3em]{MultiHate} 
 & Hateful & Represent the text based on whether it is hateful or not. \\
\cmidrule{2-3}
 & Language & Represent the text based on the language. \\
\midrule
Big Patent 
 & Patent Category & Represent the text based on the category the patent belongs to according to the Cooperative Patent Classification (CPC) code. \\
\midrule
\multirow{2}{*}[-0.3em]{IntentEmo} 
 & Intent & Represent the intent of this text. \\
\cmidrule{2-3}
 & Emotion & Represent the emotion of this text. \\
\midrule
Toxic 
 & Toxic & Represent the text based on whether it is toxic (yes/no). \\
\midrule
AG-News 
 & Topic & Represent the news according to their topic category. \\
\midrule
\multirow{2}{*}[-0.3em]{NYTClust} 
 & Topic & Represent the text based on the main news category. \\
\cmidrule{2-3}
 & Location & Represent the text based on where the news happen. \\
\bottomrule
\end{tabular*}
\end{table*}

\twocolumn

\end{document}